\documentclass[letterpaper]{article} 
\usepackage{times}  
\usepackage{helvet}  
\usepackage{courier}  
\usepackage[hyphens]{url}  
\usepackage{graphicx} 
\usepackage{amssymb}
\usepackage{booktabs}
\usepackage{natbib}  
\usepackage{caption} 
\usepackage{aaai24}
\usepackage{comment}
\usepackage{algorithm}
\usepackage{algorithmic}
\usepackage{amsmath}

\usepackage{subcaption}
\usepackage{newfloat}
\usepackage{listings}
\DeclareCaptionStyle{ruled}{labelfont=normalfont,labelsep=colon,strut=off} 
\floatstyle{ruled}
\newfloat{listing}{tb}{lst}{}
\floatname{listing}{Listing}
\title{\textit{UrzaGPT}: LoRA-Tuned Large Language Models for \\ Card Selection in Collectible Card Games}
\author{
    Timo Bertram
}
\affiliations{Institute for Machine Learning\\
    Johannes Kepler University\\
    Linz, Austria \\
    bertram@ml.jku.at
}

\begin{document}

\maketitle

\begin{abstract}

Collectible card games (CCGs) are a difficult genre for AI due to their partial observability, long-term decision-making, and evolving card sets. Due to this, current AI models perform vastly worse than human players at CCG tasks such as deckbuilding and gameplay. In this work, we introduce \textit{UrzaGPT}, a domain-adapted large language model that recommends real-time drafting decisions in \textit{Magic: The Gathering}. Starting from an open-weight LLM, we use Low-Rank Adaptation fine-tuning on a dataset of annotated draft logs. With this, we leverage the language modeling capabilities of LLM, and can quickly adapt to different expansions of the game.  We benchmark \textit{UrzaGPT} in comparison to zero-shot LLMs and the state-of-the-art domain-specific model. Untuned, small LLMs like Llama-3-8B are completely unable to draft, but the larger GPT-4o achieves a zero-shot performance of 43\%. Using \textit{UrzaGPT} to fine-tune smaller models, we achieve an accuracy of 66.2\% using only 10,000 steps. Despite this not reaching the capability of domain-specific models, we show that solely using LLMs to draft is possible and conclude that using LLMs can enable performant, general, and update-friendly drafting AIs in the future.

\end{abstract}

\section{Introduction}

For decades, novel developments in artificial intelligence have been tested on games. Games provide structure, and games allow for easy comparison, but games can also be incredibly difficult. In the past, abstract board games were the prime target for AI \cite{brown2019SuperhumanAIMultiplayer, silver2018GeneralReinforcementLearning}. More recently, video games and strategy games also have been explored as test environments \cite{vinyals2019GrandmasterLevelStarCraft, openai2019Dota2Large}. Modern games are varied in their rules, which can even change in expansions or updates. Video games require many actions per second, fine-grained control, and efficient image processing \cite{pearce2022CounterStrikeDeathmatchLargeScale, vinyals2019GrandmasterLevelStarCraft}. Strategy games often require language processing and long-horizon planning \cite{hoover2020ManyAIChallenges}. Despite AI having achieved expert-level or even superhuman performance in games like chess and Go \cite{schaeffer2007CheckersSolved, schrittwieser2020MasteringAtariGo}, only weak game agents exist for many contemporary games. Collectible card games (CCGs) are one of those genres for which currently, no agents exist that perform on the level of human experts.

\begin{figure}[htbp]
    \centering
    \begin{subfigure}[b]{0.45\columnwidth}
        \includegraphics[width=\textwidth]{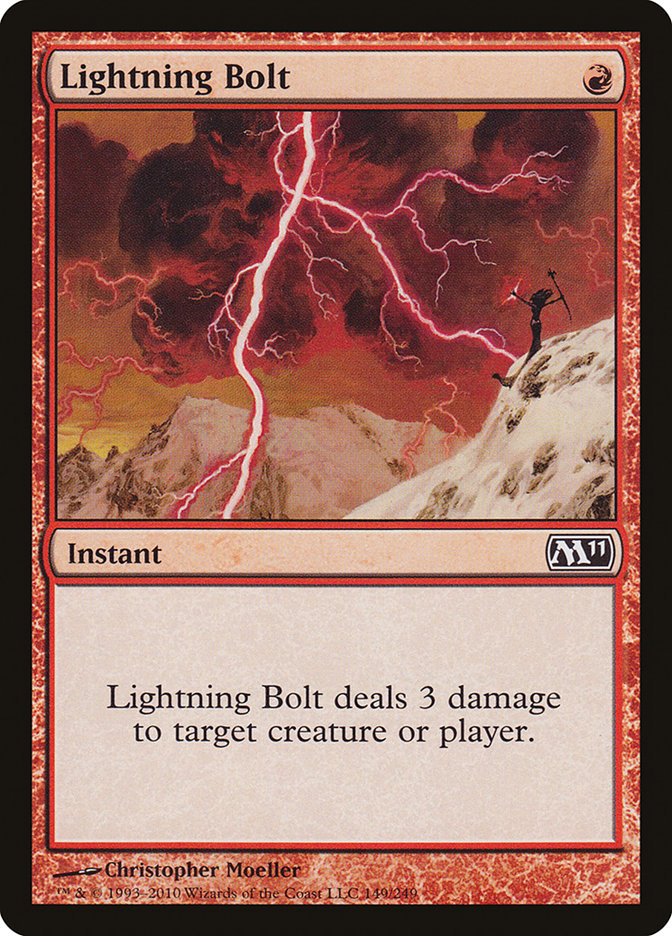}
        \label{fig:bolt}
    \end{subfigure}
    \hfill
    \begin{subfigure}[b]{0.45\columnwidth}
        \includegraphics[width=\textwidth]{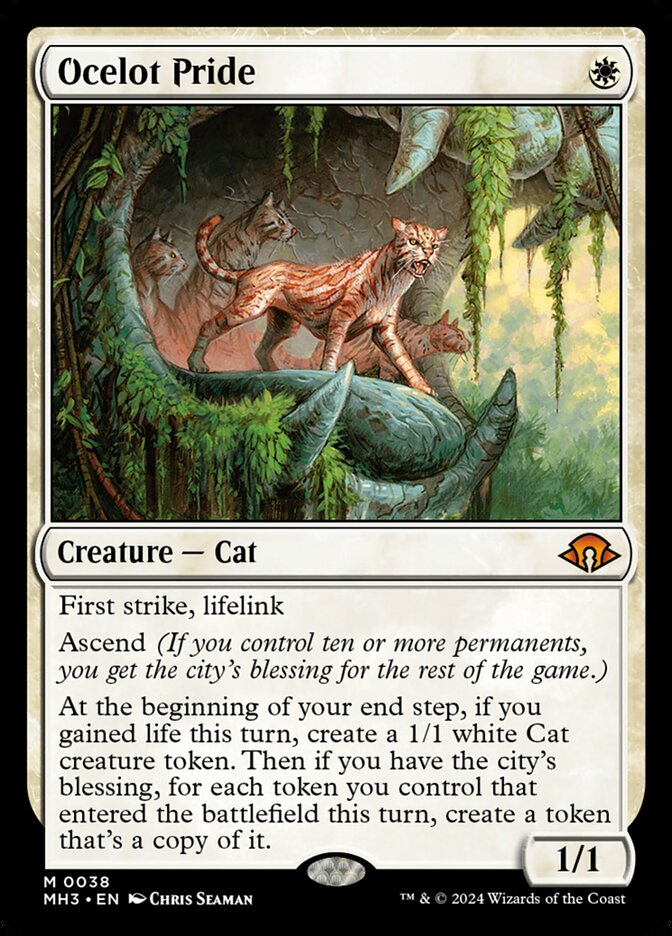}
        \label{fig:ocelot}
    \end{subfigure}
    \caption{\textit{Magic: The Gathering} cards come in various levels of complexity, which makes it difficult to process them in a model. While it is straightforward to numerically represent the card on the left, representing the card on the right requires a semantic understanding of the game-specific language. Therefore, large language models are a natural fit to evaluate collectible cards.}
    \label{fig:card_comparison}
\end{figure}

The recent explosion of language modelling due to transformers \cite{vaswani2017AttentionAllYou} and large-scale pre-training \cite{radford2018ImprovingLanguageUnderstanding} may provide a solution for this. Large Language Models (LLMs) compress a vast amount of information into one model and thus learn about games and strategy. LLMs are inherently language-based, which makes it simple to describe rules and abilities. This is especially crucial for CCGs, where a large part of their complexity comes from the natural language used to describe the functions of cards (see Figure~\ref{fig:card_comparison}).

In this work, we introduce \textit{UrzaGPT}, a domain-adapted LLM for drafting \textit{Magic: The Gathering}, and explore its utility. Specifically, we contribute the following:

\begin{itemize}
    \item An evaluation of the zero-shot drafting performance of popular LLMs.
    \item We fine-tune LLMs to drafting \textit{Magic: The Gathering} and show how models can quickly be adapted to this domain.
    \item An ablation on the impact of fine-tuning parameters on the performance.
\end{itemize}

We begin by discussing related work (Section~\ref{sec:related_work}) and briefly explaining the most critical game concepts (Section~\ref{sec:MTG}) before outlining the specifics of our method in Section~\ref{sec:method} and discussing the results in Section~\ref{sec:results}. All code will be released upon acceptance.

\begin{figure*}
    \centering
    \includegraphics[width=\linewidth]{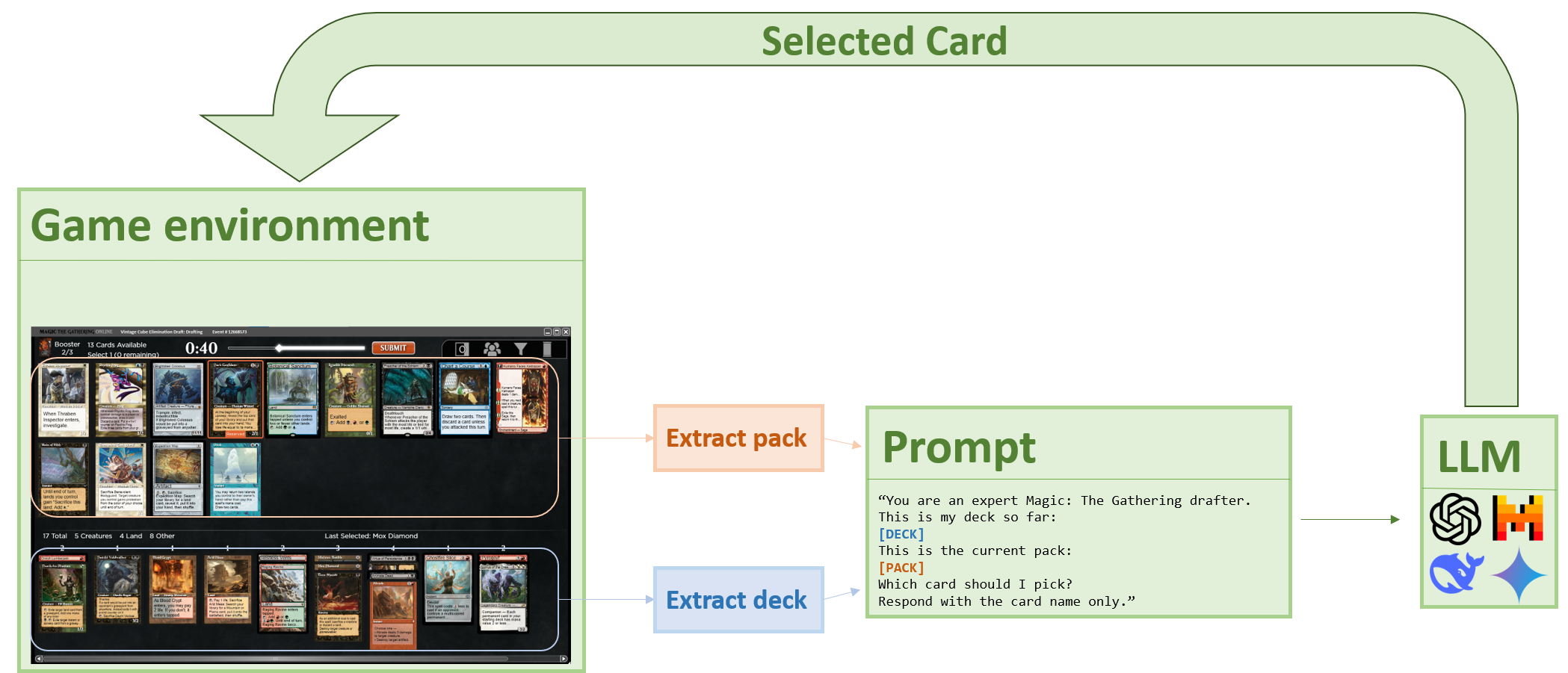}
    \caption{High-level overview of the \textit{UrzaGPT} pipeline. Pack and deck information is extracted from the game, cards are transformed into a natural language representation, and an LLM is prompted with this information. The LLM responds with a single card name, which is fed back into the game environment to take an action.}
    \label{fig:overview}
\end{figure*}

\section{Related Work}
\label{sec:related_work}
\noindent\textbf{Artificial Intelligence in Competitive Games.}
In the past, artificial intelligence for games was often restricted to structured board and card games such as chess \cite{silver2018GeneralReinforcementLearning, campbell2002DeepBlue}, Go \cite{silver2016MasteringGameGo, silver2017MasteringGameGo}, checkers \cite{schaeffer2007CheckersSolved}, or poker \cite{moravcik2017DeepStackExpertlevelArtificial, brown2019SuperhumanAIMultiplayer}. This extended to faster, commercial games like Doom \cite{lample2017PlayingFPSGames}, Starcraft II \cite{vinyals2019GrandmasterLevelStarCraft}, Dota 2 \cite{openai2019Dota2Large}, and Gran Turismo \cite{wurman2022OutracingChampionGran}. The majority of these use some variant of reinforcement learning to train from scratch, requiring no human data but fast simulation environments and a large amount of computation. The majority of agents are completely domain-specific, offering no way to adapt to a similar environment and require complete retraining on a new task. \newline

\noindent\textbf{LLM-based GameAI.}
The recent advancements in natural language processing have led to a large interest in using language models as game agents \cite{hu2025SurveyLargeLanguage, gallotta2024LargeLanguageModels, sweetser2024LargeLanguageModels, yang2024gpt}. Game environments have been used to benchmark LLM and VLM reasoning \cite{paglieri2025BALROGBenchmarkingAgentic, costarelli2024GameBenchEvaluatingStrategic}, and \textit{TextStarcraft II} adapts the real-time strategy game into a text format for LLMs \cite{ma2024LargeLanguageModels}. Additionally, work in LLMs for games has tackled Pokémon \cite{hu2024PokeLLMonHumanParityAgent}, Slay the Spire \cite{bateni2024LanguageDrivenPlayLarge}, Werewolf \cite{xu2024ExploringLargeLanguage}, and Minecraft \cite{zhu2023GhostMinecraftGenerally, wang2024DescribeExplainPlan}. Despite all of this work, the performance of LLM-based agents is often subpar when not making further adjustments to their actions. For example, modern LLMs still fail to play TicTacToe perfectly, with the exact prompt used and model choice having a large influence \cite{topsakal2024BenchmarkingLargeLanguage}. In the BALROG benchmark, many models achieve some progress on the easier games, but more complicated environments like NetHack \cite{kuttler2020NetHackLearningEnvironment} are currently out of reach. \newline

\noindent\textbf{Drafting and Deckbuilding in Collectible Card Games.} Research on game agents for collectible card games is mostly focused on using reinforcement learning or genetic algorithms \cite{dockhorn2019IntroducingHearthstoneAICompetitiona, kowalski2023SummarizingStrategyCard}. In both cases, heuristic agents are used to evaluate decks, while decks are either created sequentially \cite{esilvavieiraExploringReinforcementLearning2023, yang2021DeckBuildingCollectible} or evolved \cite{kowalski2020EvolutionaryApproachCollectible, garcia-sanchez2016EvolutionaryDeckbuildingHearthstone}. When a fast simulator exists, a playing policy can be learned end-to-end with deckbuilding, which decouples the agent from the playing heuristics \cite{xi2023MasteringStrategyCard, xiao2023MasteringStrategyCard}. Lastly, some work directly learns card selection from human data \cite{bertram2021PredictingHumanCard, ward2021AISolutionsDrafting}. The current state-of-the-art in drafting prediction uses a contrastive model that projects cards and decks into a learned space to make selections \cite{bertram2024LearningGeneralisedCard}.

Our work focuses on drafting as a pure next-token prediction task. Previous work uses language embeddings to represent cards but still trains a classification model from scratch \cite{bertram2024LearningGeneralisedCard}. We skip the classification model and directly fine-tune an LLM to predict the chosen cards.

\section{Magic: The Gathering}
\label{sec:MTG}
Collectible card games like \textit{Magic: The Gathering}, \textit{Hearthstone}, or \textit{Lorcana}, are competitive, turn-taking games. As card games, they feature a large amount of stochasticity and imperfect information. Additionally, they use thousands of unique, highly complex cards, which makes representation and adaptation difficult. In \textit{Magic: The Gathering}, the goal of gameplay typically is to reduce the opponent's life points to zero. To achieve that, players have to first build their decks, i.e., decide which cards they want to play with, and later use the chosen cards to win the game. This reveals two connected, but also distinct, challenges of the game: \textbf{deckbuilding} and \textbf{gameplay}. Deckbuilding is concerned with which cards to use, while gameplay regards how to use them. In this work, we solely focus on deckbuilding and disregard the gameplay. However, LLMs might also enable gameplay in the future.

\subsection{Drafting}
Deckbuilding in \textit{Magic: The Gathering} exists in different forms. Before the game, players decide the format they want to use:

\noindent\textbf{Constructed} allows players to freely choose from all available cards. Decks have to contain at least 60 cards, and every card can be put into a deck up to 4 times. Which cards are available depends on the chosen restrictions, but the total number of unique cards to choose from currently ranges from 4,095 when playing \textit{Standard} to 29,084 in \textit{Vintage}.\footnote{According to \url{https://scryfall.com/search?q=legal\%3Astandard} and \url{https://scryfall.com/search?q=legal\%3Avintage}} Therefore, a lower bound of potential deck configurations in \textit{Standard} and \textit{Vintage} is given by $\binom {4095 \cdot 4}{60} >10^{170}$ and $\binom {29084 \cdot 4}{60} > 10^{222}$ respectively.

\noindent\textbf{Drafting} significantly restricts the deckbuilding process by turning it into a sequential process. Instead of freely choosing from a large pool, players are repeatedly tasked with selecting a single card out of a \textit{pack} of initially 15 pseudo-randomly distributed cards. Players simultaneously select one card out of the 15 and, after choosing one, pass all unchosen cards to the player next to them, who has to select from the 14 remaining cards. Thus, throughout a draft, the number of possible options reduces until all 15 cards are chosen and the whole process is repeated twice more, leading to a total of 45 per player. As a side note, drafts are typically played with 8 players. This means that every player sees every pack of cards \textbf{twice}, once with $n$ cards and once with $n-8$ cards, from which small amounts of information about the opponents' preferences can be deduced. Most of the time, players draft with packs from a single \textit{expansion} of the game, which means that only about 200-300 unique cards can appear in packs. Expansions are highly unique, with new expansions often containing mostly completely new cards. This makes it difficult to adapt to the new expansion. 

In this work, we only regard drafting here and do not yet aim to model constructed deckbuilding. 

\subsection{Colors}
\label{sec:colors}
While we will not explain the rules of \textit{Magic: The Gathering} in detail here, we briefly want to explain a concept that is important for later explanations; \textbf{colors}. \footnote{We make some simplifications here that are not completely correct for all cards.} In MTG, cards either require mana (\textit{spells}) or produce mana (\textit{lands}). Spells and mana are colored, i.e., red spells will require red mana produced by a \textit{Mountain}. Due to this restriction, it is possible to draw combinations of lands and spells that do work together, i.e., red mana and blue spells. Therefore, most decks will only contain 1 to 2 out of the 5 colors in the game. This provides an easy baseline for sensible drafting: if a model frequently picks cards not in the colors of the deck, it does not understand drafting. We will later refer back to this concept and investigate whether LLMs adhere to it.

\section{Method}
\label{sec:method}
Our hypothesis is straightforward: LLMs excel at processing language, and \textit{Magic: The Gathering} cards are largely language-based, so LLMs should be able to evaluate and draft cards. Since there are no fast simulators for the full game available, it is not possible to test the drafted decks at scale. Rather, we resort to human data as the ground truth and check whether the LLMs' actions line up with the human players'.

\subsection{Data}
We use open-access datasets available at \url{17lands.com}. The datasets contain step-by-step drafts of human players, which allow us to reconstruct which card was picked at every decision point. Data is collected solely from users of the \url{17lands.com} website who downloaded a tool that records their games. This in itself selects for a subset of proficient players, who are invested enough into the game such that they would go through this process. Additionally, data contains information about the rating of players, allowing for additional refinement to high-ranked players if necessary.

We preprocess the datasets by transforming them into a language-only format, noting the cards currently in the player's deck, the cards in the pack from which the player was allowed to choose, and the chosen card. From this, the task is to predict the chosen card, given the pack and the deck. Note that this is a noisy problem, as there is no objectively correct answer, but rather a prediction of human decisions. It is possible that another player would make a different decision, but we simply regard the chosen card as the ground truth.

There are multiple datasets with different \textit{expansions} of the game available, but we restrict this study to the \textit{NEO} dataset. The \textit{NEO} dataset contains 5,122,921 pick decisions, although we find that we did not require all of them. We used a subset of 10,000 samples as the test set for later experiments. The data contains a mixture of drafts from \textit{Premier draft} and \textit{Traditional draft}, the majority of which is \textit{Premier}. These two variations have slight differences, e.g., in \textit{Traditional}, sideboard cards can become important, but we regard them all the same. The data contains information about the players' proficiency, such as their rank and results, but we found no significant difference in results when filtering for high-ranked players.

\subsection{Card Representation}

There are two ways to describe cards in the prompt for the LLMs. One can either simply use the cards' names, or one can use the full card text.

\noindent\textbf{Names} are an easy identifier and vastly reduce the size of the prompt, which leads to increased throughput. However, only using card names makes the strong assumption that models have seen plenty of data on these cards in pre-training, such that they are able to associate a card's name with their abilities. Including the full card \textbf{text} with all values and categories on the card seems intuitively more useful, as it provides more information, but there are issues with it. Firstly, card texts are long and thus the prompt grows immensely compared to just using card names. In our experiments, a single card required ~100-300 tokens. With up to 44 cards in the deck and packs, we reach the maximum context length for some standard models, e.g., Llama-3-8B's maximum of 8,000 tokens. In addition, it has been shown that large context sizes can lead to information retrieval issues \cite{machlab2024LLMInContextRecall}. In the following section, we provide a small ablation with both versions and compare their performance.

\subsection{LoRA Fine-Tuning}

While large language models (LLMs) possess extensive general knowledge, they are not optimized for specific downstream tasks. Fine-tuning is a common approach to adapt these models to new domains or tasks. In full fine-tuning, all model parameters are updated using a task-specific dataset. Although this can achieve strong performance, it requires significant computational resources and storage. To address these limitations, \textit{parameter-efficient fine-tuning (PEFT)} techniques have been proposed, which reduce the number of trainable parameters while maintaining performance.

In this work, we use \textit{Low-Rank Adaptation (LoRA)}~\cite{hu2022lora} to fine-tune LLMs. LoRA freezes the original model weights and introduces trainable \textit{low-rank matrices} into the attention layers. Specifically, given a weight matrix $W \in \mathbb{R}^{d \times k}$ in the transformer (e.g., a query or value projection matrix), LoRA reparameterizes it as:

\[
W' = W + \Delta W = W + AB
\]

where $A \in \mathbb{R}^{d \times r}$ and $B \in \mathbb{R}^{r \times k}$ are the low-rank matrices with rank $r \ll \min(d, k)$, and only $A$ and $B$ are trained during fine-tuning. This approach significantly reduces the number of trainable parameters and memory footprint.

Importantly, LoRA introduces these updates in a way that maintains the original model's pretraining knowledge, and it can be implemented without modifying the model architecture. It has been shown to reduce the number of trainable parameters by up to 10,000$\times$ compared to full fine-tuning whileachieving comparable performance on many tasks.

\section{Experiments}
We begin by checking the zero-shot performance of popular LLMs without using domain-specific data. Then, we move towards fine-tuning them to improve upon that baseline. All experiments use the same prompt (Appendix~\ref{app:prompt}). The prompt was iterated on several times in the zero-shot setting until we achieved decent results. Notably, the prompt only regards the current decision without a history of previous pick options, thus omitting signals that players can pick up on while drafting. This decision was made to allow for a clear comparison of architectures to prior work.

\subsection{Zero-Shot Drafting with LLMs}
For the zero-shot drafting experiment, we selected multiple models of similar parameter sizes (7 to 8 billion) and queried them for drafting selections out of the box. 

\begin{figure}[ht]
    \centering
    \includegraphics[width=\columnwidth]{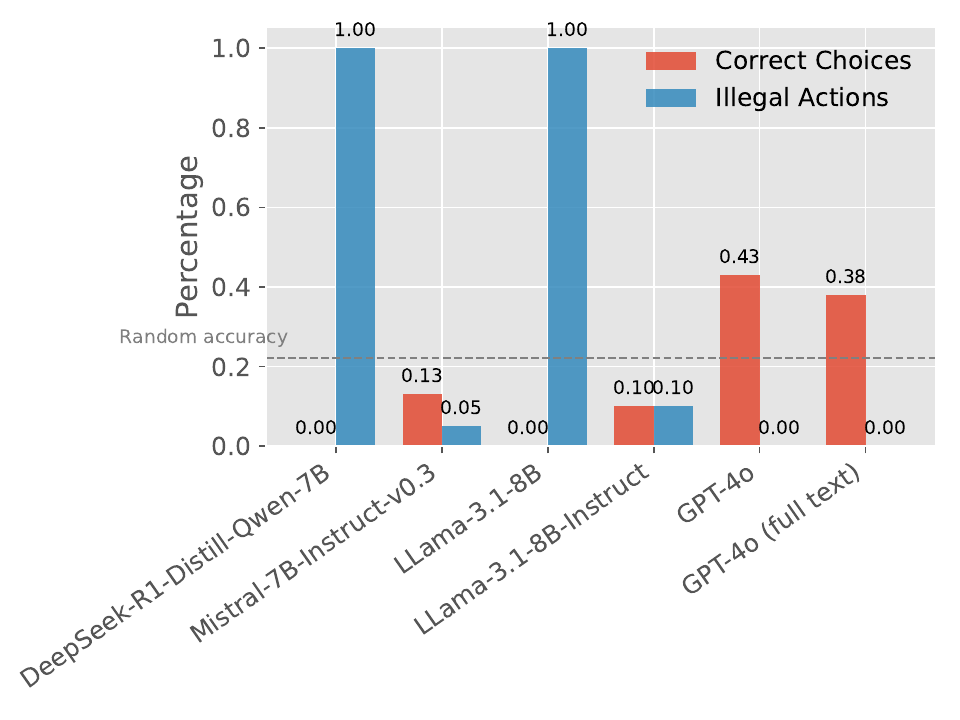}
    \caption{Zero-shot selection accuracy and illegal selections of different models. A correct selection is counted when the output clearly selects the true chosen card in the test data. Illegal selections are when the model provides an unclear output or selects a card that was not offered. The larger GPT-4o models perform vastly better than the smaller and distilled models. We also find that using the full card text slightly harms performance (43\% vs 38\%).}
    \label{fig:zero_shot}
\end{figure}

Our prompt includes ``Respond with the card name only." but we did not negatively score models that did not adhere to it. We found that many models provided long outputs going through iterative thinking processes, so we manually evaluated whether the final decision of the model lined up with the truth in the data. Because this experiment required manual extraction of the card selection, we restricted it to 100 samples per model. Figure ~\ref{fig:zero_shot} shows the result of this.

We see that the smaller 7-8 billion parameter models are unable to follow the prompt, which leads to illegal actions. Illegal actions are defined as either not making a concrete selection in the response of the model or by picking a card that was not available for selection. However, the large GPT-4o models both follow the prompt strictly, always answering with the card name only and always choosing a possible card. Although both of these perform better than the random baseline of 0.221, we find that including the complete card text in the prompt slightly decreases the accuracy from 0.43 to 0.38. This suggests that the models already possess knowledge of the cards, that only using card names provides enough information, and that including the full text dilutes the prompt. This is validated with knowledge prompts (Appendix~\ref{app:knowledge_questions}), which show that the small Llama model is unable to understand the card and makes errors in the description of the drafting process, while the large GPT-4o perfectly replicates the ability of the card.

\subsection{Fine-Tuning}
The results from the zero-shot experiment show promising results for the large models, but the 7B and 8B models are unable to draft. In this experiment, we investigate how supervised fine-tuning influences the performance.

For this, we train the \textit{Mistral-7B-Instruct} and the \textit{Llama-3-8B} \cite{touvron2023Llama} models on the aforementioned datasets. We train on 1 million picks and evaluate on a held-out test set.

\noindent\textbf{Setup} We train the models on prompt completion, i.e., we give in the complete prompt (Figure~\ref{fig:overview}) and use the cards in the dataset as the target completion. To make fine-tuning more computationally efficient, we use LoRA \cite{hu2022lora} with a rank of 8 and an alpha of 16 (Appendix~\ref{app:hyperparams}). For training, we use a batch size of 8 with 4 gradient accumulation steps, leading to an effective batch size of 32. Training was done on 3 Nvidia A100 GPUs and took 6 hours per model.

\begin{figure}
    \centering
    \includegraphics[width=\columnwidth]{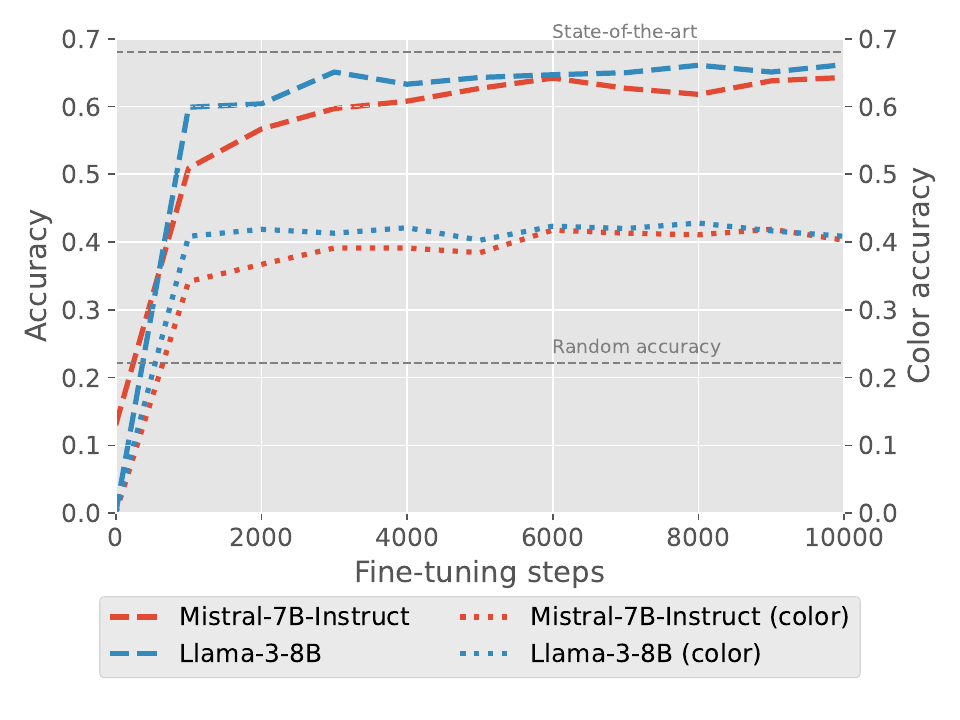}
    \caption{Accuracy of predicted selections of the \textit{Llama-3-8b} and \textit{Mistral-7B-Instruct} models when supervised fine-tuning. Accuracy quickly rises at the start and makes further improvements up to a peak of 66.2\% after 10,000 steps for \textit{Llama-3-8B}. \textit{Mistral-7B-Instruct} reaches a lower peak accuracy of 64.3\%. This performance almost reaches the accuracy of domain-specific models.}
    \label{fig:accuracy_over_time}
\end{figure}

We measure two metrics; the accuracy of predictions and the adherence of the model to the primary colors of the deck (see Section~\ref{sec:colors}). For this, we simply check whether the chosen card's color is in the two primary colors of the current deck, as given by the sum of cards of each color. This is a slightly inaccurate metric, due to staying on color not always being the correct choice, but nonetheless gives insight into the decisions. 

We find that fine-tuning quickly increases the accuracy of the models, reaching an accuracy of 60\% after roughly 1,000 steps (see Figure~\ref{fig:accuracy_over_time}). Further training increases the accuracy more, with \textit{Llama-3-8B} reaching a peak accuracy of 66.2\%. \textit{Mistral-7B-Instruct} performs slightly worse with an accuracy of 64.3\%. 

Although 66.2\% is a vastly higher accuracy than the initial zero-shot performance of 13\%, and outperforms the zero-shot performance of the large GPT-4o models (43\%), this result still falls short of the performance of domain-specific drafting models (68\%). Despite this, the gap between the fine-tuned LLMs and domain-specific models is small.

\subsection{Influence of Rank on Performance}
We validate how the chosen parameters of the fine-tuning process influence the resulting models. Most crucially, the \textit{rank} of the LoRA adapters controls how many parameters can be changed in the fine-tuning process, offering better adaptability when using more computation.

\begin{figure}
    \centering
    \includegraphics[width=\columnwidth]{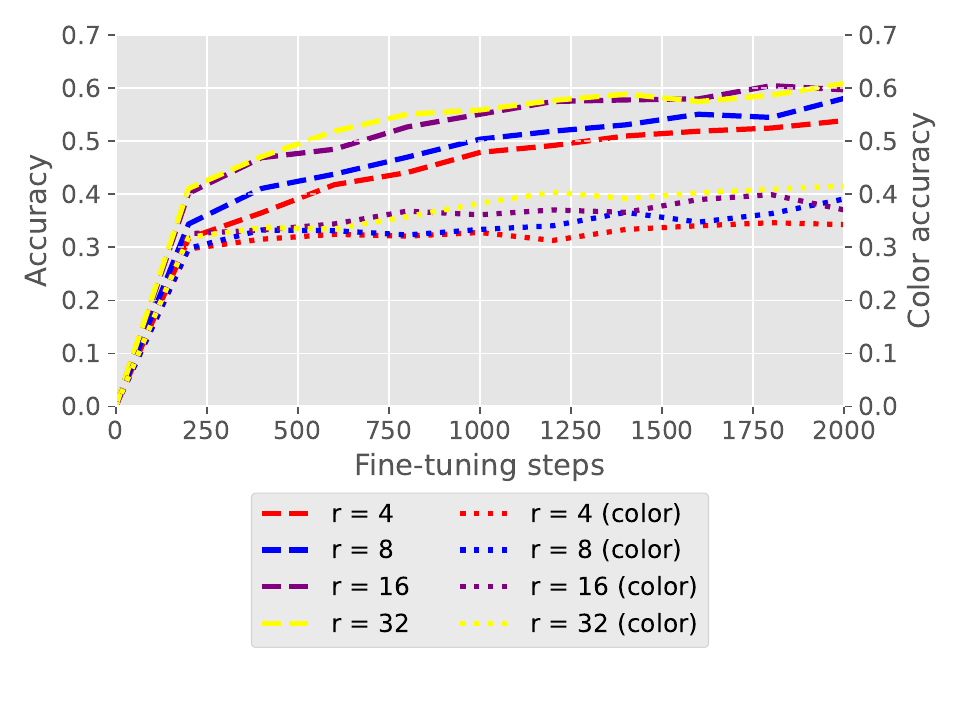}
    \caption{The influence of LoRA rank on the performance of \textit{Llama-3-8B}. We find that increasing the rank yields better performance, both in accuracy of predictions and adherence to the color of the deck, but diminishing returns kick in at a rank of 16. This shows a clear trade-off between the parameter count of the fine-tuning process and the resulting performance.}
    \label{fig:accuracy_rank}
\end{figure}

Figure~\ref{fig:accuracy_rank} shows the results for \textit{Llama-3-8B}. As expected, using more parameters to adapt the model yields better performance. We also see that both the total accuracy and the color accuracy are influenced. This lets us believe that using more parameters allows for a better understanding of the domain and remembering the color of cards.

\section{Discussion}

Our results show that LLMs possess the capabilities to select cards in a drafting context with minimal exposure to domain-specific data. However, several important observations and open questions arise from our findings.

\textbf{Zero-shot capabilities.} The performance gap between small and large LLMs in the zero-shot setting suggests that model scale plays a crucial role in encoding niche domains like \textit{Magic: The Gathering}. While GPT-4o achieves reasonable zero-shot performance (43\% accuracy), smaller models like Llama-3-8B struggle without additional tuning. This highlights that mere exposure to natural language isn't sufficient to learn drafting heuristics unless the model has seen significant domain-relevant data.

\textbf{Data efficiency of fine-tuning.} Remarkably, we find that with just 1,000 steps of LoRA fine-tuning, models can exceed 60\% accuracy—outperforming larger, untuned models. This demonstrates strong transfer potential and the utility of supervised data, especially for structured decision-making tasks. However, performance plateaus rather quickly, indicating a ceiling imposed either by model capacity or the limits of the training objective.

\textbf{Representation trade-offs.} Including the full card text in prompts surprisingly reduced performance in the zero-shot setting. This suggests that longer contexts introduce unnecessary noise or exceed the model's ability to retrieve key information. It may also reflect pretraining limitations, where the model's understanding of card names is learnt but full-text parsing of novel cards is not.

\textbf{Limitations of accuracy metrics.} While accuracy is a straightforward measure for predicting card picks, it fails to capture the quality of full drafts or synergy across picks. Two picks may be equally valid depending on downstream deck strategy and personal preferences influence decisions. Thus, future work should consider reinforcement learning from outcomes or incorporate gameplay simulations to better evaluate utility.

\textbf{Transferability and generalisation.} Our study focused on a single expansion (NEO), leaving open the question of generalisation. Can models trained on one set adapt to others with similar mechanics? Are their representations reusable across expansions? NEO was chosen to compare to prior work, but is ultimatively an arbitrary choice, thus performance should translate to other sets, but more work is required to investigate this in depth.

\textbf{LoRA parameter count.} We find that introducing more trainable parameters has a positive correlation with the fine-tuned model's performance. Thus, an open question is how far one can push this performance. Would more parameters increase accuracy further? Would full fine-tuning exceed the PEFT setting? More work is required to investigate this in depth.

Overall, our findings support the viability of LLM-based drafting agents, but also reveal limitations in LLMs' general capability for highly strategic tasks.

\section{Conclusion}
\label{sec:results}

This work introduces \textit{UrzaGPT} and takes the first steps towards fully LLM-based drafting in the collectible card game \textit{Magic: The Gathering}. We frame drafting, i.e., selecting cards, as a pure language task, prompting an LLM with the current cards in the deck and the available options. 

Out of the box, smaller LLMs (7 or 8 billion parameters) do not perform well on this task. They largely do not adhere to the prompt and due to this perform worse than random. Large LLMs (GPT-4o) perform much better and achieve a 43\% zero-shot accuracy. This clearly shows the general game-playing capabilities of LLMs, which seem to have been trained on \textit{Magic: The Gathering} data. GPT-4o is able to draft solely from card names. However, we find that including the cards' abilities in the prompt diminishes performance, likely due to diluting the prompt. 

Small amounts of fine-tuning allow even the 7 billion parameter models to draft competitively. Fine-tuning the whole model is unfeasible even for the smaller models, so we use LoRA to adapt the models to the task. After only 1,000 steps with LoRA-tuning, the LLMs reach an accuracy of 60\%, only 8 percentage points behind the state-of-the-art drafting model, underlining the potential of LLMs for gameplaying tasks. Further training yields increasingly better performance, but the LLMs show a lower ceiling than that of domain-specific models (66.2\% vs 68\%). 

Altogether, we show that quick adaptation of small-scale LLMs to drafting is possible. In the future, this can enable low-cost agents and player aids that can quickly adapt to new expansions of the game.

\section{Limitations and Future Work}
This work is limited by its reliance on historical data. We assume that cards selected by players form an objective ground truth, despite this not being generally true. Rather than regarding each pick as an independent decision, it would be more truthful to the domain to evaluate the final deck after a complete draft. In the future, we aim to bring \textit{UrzaGPT} to the real world, using simulators to validate the final decks and to plug into real players' drafts. 

We also find that larger models provide better results. These models might not be computationally feasible without large resources.

\section{Ethical Statement}

The existence of a super-human drafting agent could have an impact on the online versions of the game as it provides an advantage to players who use them. However, drafting companions already exist and are an accepted tool. Addtionally, current models are not on a human level and should only be used as auxiliary helpers and should not be followed without validation.

\appendix
\section{Hyperparameters}
\label{app:hyperparams}
\begin{table}[H]
\label{tab:hyperparams}
\centering
\begin{tabular}{cc}
\toprule
Parameter & Value\\
\midrule
Maximum sequence length & 512\\
Maximum token output & 30 \\
\midrule
Num Devices & 3\\
Device Batch-Size & 8\\
Gradient Accumulation Steps & 4\\
\midrule
LoRA Rank & 8\\
LoRA Alpha & 16\\
LoRA Dropout & 0.05\\
\end{tabular}
\end{table}

\section{Prompt}
\label{app:prompt}
\textit{
    ``You are an expert Magic: The Gathering drafter.
    My pool so far:
    [LIST OF POOL OF CARDS]
    Current pack:
    [LIST OF CARDS IN PACK]
    Which card should I pick? Respond with the card name only."
}

\section{LLM Knowledge Questions}
\label{app:knowledge_questions}

Here we ask some general \textit{Magic: The Gathering} and drafting questions to pre-trained models. This serves to underpin their selections and investigate whether they know cards well.

\subsection{Q1: How does drafting work in Magic: The Gathering and what do you have to pay attention to?}
\noindent\textbf{GPT-4o:} Great question! Drafting is one of the most skill-testing and fun ways to play Magic: The Gathering. It combines both deckbuilding and gameplay, and it's a popular format for tournaments and casual play alike. Here's how it works and what you should pay attention to:

\textbf{How Drafting Works (Booster Draft Format)}
\begin{enumerate}
    \item Players \& Packs:
    \begin{itemize}
        \item Usually 8 players sit in a pod.
        \item Each player gets 3 booster packs (often from the same set or compatible sets).
    \end{itemize}
    \item The Draft Process:
    \begin{itemize}
        \item Everyone opens their first pack, picks one card, then passes the remaining cards to the left.
        \item You receive a pack from the player to your right, select a card, and repeat.
        \item Once all cards are drafted from the first pack, open the second pack, but pass to the right this time.
        \item Third pack passes to the left again.
    \end{itemize}
    \item Deck Construction:
    \begin{itemize}
        \item After drafting, build a minimum 40-card deck, including lands.
        \item Typically, you’ll have around 23 spells and 17 lands.
        \item You can use as many basic lands as you want, provided by the event organizer or store.
    \end{itemize}
\end{enumerate}

\textbf{Key Things to Pay Attention To}
\begin{enumerate}
    \item Signaling and Reading the Table
    \begin{itemize}
        \item Watch what colors or strategies seem open.
        \item If strong cards of a color keep coming late, that color is likely under-drafted.
        \item Conversely, if the color you’re drafting dries up early, others at the table may be in the same color.
    \end{itemize}
    \item Color Commitment
    \begin{itemize}
        \item Ideally, settle into two main colors, but stay flexible early on.
        \item Don't over-commit in the first few picks unless you open a powerful ``bomb" worth building around.
    \end{itemize}
    \item Card Evaluation
    \begin{itemize}
        \item Prioritize:
        \begin{itemize}
            \item \textbf{Bombs:} Game-winning cards, especially rares or mythics.
            \item \textbf{Efficient Removal:} Cards that can deal with your opponent’s threats.
            \item \textbf{Efficient Creatures:} Good stats for cost, especially with abilities like Evasion (Flying, Menace).
        \end{itemize}
        \item Lower priority:
        \begin{itemize}
            \item Situational cards or narrow build-around cards (unless your deck supports them).
        \end{itemize}
    \end{itemize}
    \item Curve Considerations
    \begin{itemize}
        \item Pay attention to your mana curve:
        \begin{itemize}
            \item Early drops (1-2 mana creatures) are important for pressure.
            \item Midrange creatures (3-5 mana) fill your deck's core.
            \item A few high-end bombs or big creatures can finish games.
        \end{itemize}
    \end{itemize}
    \item Synergy vs Raw Power
    \begin{itemize}
        \item Sometimes synergy (like tribal interactions or archetype-specific combos) outweighs raw card quality.
        \item Look for set-specific archetypes (e.g., ``Blue-Red Spells" or ``Black-Green Graveyard" themes) and draft accordingly.
    \end{itemize}
    \item Fixing and Splashing
    \begin{itemize}
        \item Include mana-fixing if you plan to splash a third color for powerful cards.
        \item Only splash if it's worth it and your mana can support it.
    \end{itemize}
\end{enumerate}

\textbf{Additional Tips}
\begin{itemize}
    \item \textbf{Sideboard cards:} Draft situational cards like artifact hate or graveyard interaction—they go into your sideboard for specific matchups.
    \item \textbf{Draft signals:} Pay attention to what cards you pass—good players will ``read" that information.
    \item \textbf{Set Knowledge:} Familiarity with the set's mechanics, archetypes, and common tricks gives you an edge.
\end{itemize}

\textbf{Example:}
In a typical draft of Wilds of Eldraine, you might look for synergies like ``Faeries" (Blue-Black) or ``Food tokens" (Green-Black) and value cards accordingly.

If you want, I can walk you through a simulated draft or analyze pick orders for a specific set. Want to dive deeper?\\

\noindent\textbf{Llama-3-8B-Instruct:} Drafting in Magic: The Gathering is a format where players create a deck from a pool of randomly selected cards. Here's a comprehensive guide to help you understand how drafting works and what to pay attention to:

\textbf{Basic Drafting Rules}
\begin{enumerate}
    \item \textbf{Sealed Pool:} Before the draft, each player receives a sealed pool of 15-20 cards, which includes a random selection of cards from a specific set or sets.
    \item \textbf{Drafting:} Players take turns selecting cards from their pool to add to their deck. Each player has a set amount of time to make their picks.
    \item \textbf{Deck Building:} Players can only add one card to their deck at a time. They can remove cards from their pool, but they cannot add new cards to their pool once the draft is complete.
    \item \textbf{Deck Size:} A standard draft deck consists of 40 cards.
\end{enumerate}

\textbf{Drafting Strategy}

To succeed in drafting, you need to pay attention to several factors:

\begin{enumerate}
    \item \textbf{Mana Curve:} Balance the number of low-cost, mid-cost, and high-cost cards in your deck. A good mana curve helps you to consistently generate mana and play spells.
    \item \textbf{Color Identity:} Focus on building a cohesive color identity by selecting cards that fit within your chosen colors (e.g., Mono-Red, Blue-White). This helps you to play synergistic cards and develop a consistent strategy.
    \item \textbf{Card Advantage:} Prioritize cards that provide card advantage, such as card draw, removal, or ramp spells.
    \item \textbf{Card Interaction:} Select cards that interact well with each other, such as synergistic combinations or counterspells.
    \item \textbf{Early Game:} Focus on building a strong early game with low-cost cards that can help you to develop your board presence.
    \item \textbf{Late Game:} Prioritize cards that provide card advantage, removal, or game-changing effects in the late game.
    \item \textbf{Board State:} Consider the board state when making picks, including the cards your opponents have played and the life totals.
    \item \textbf{Drafting for Specific Archetypes:} Familiarize yourself with popular draft archetypes and try to build around them.
\end{enumerate}

\textbf{Drafting Tips}
\begin{enumerate}
    \item \textbf{Don't Force It:} Don't feel obligated to take a card just because it's in your pool. If it doesn't fit with your strategy, it's better to pass.
    \item \textbf{Pay Attention to the Pools:} Observe the cards in your opponents' pools to anticipate their strategies and make informed picks.
    \item \textbf{Manage Your Pool:} Keep your pool organized and make sure you're not overlooking any important cards.
    \item \textbf{Communicate with Your Opponents:} Engage with your opponents and discuss the cards you're interested in or trying to avoid.
    \item \textbf{Stay Flexible:} Be prepared to adjust your strategy based on the cards you're drafting and the board state.
\end{enumerate}

\textbf{Common Drafting Mistakes}

\begin{enumerate}
    \item \textbf{Overemphasizing a Single Card:} Don't prioritize a single card over your overall strategy.
    \item \textbf{Ignoring Your Opponents' Pools:} Failing to pay attention to your opponents' pools can lead to missed opportunities and poor picks.
    \item \textbf{Not Managing Your Pool:} A disorganized pool can lead to missed cards and poor decision-making.
    \item \textbf{Not Adapting to the Draft:} Failing to adjust your strategy based on the cards you're drafting can lead to a suboptimal deck.
\end{enumerate}

By understanding the basics of drafting and paying attention to these key factors, you'll be well on your way to becoming a skilled Magic: The Gathering drafter. Happy drafting!

\subsection{Q2: What color is ``Banishing Slash" from the Magic: The Gathering" expansion ``Kamigawa: Neon Dynasty", is it strong, and when would you want to draft it for a deck?}
\noindent\textbf{GPT-4o:} Great choice! Let's break down Banishing Slash from Kamigawa: Neon Dynasty:

\textbf{Banishing Slash – Card Overview}
\textbf{Color}:
\begin{itemize}
    \item \textbf{White} (specifically a mono-White uncommon)
\end{itemize}

\textbf{Card Text:}
\textit{Banishing Slash — {W}{W}  
Sorcery  
Destroy up to one target artifact, enchantment, or tapped creature.  
Then, if you control an artifact and an enchantment, create a 2/2 white Samurai creature token with vigilance.}

\textbf{Is Banishing Slash Strong?}\\
\textbf{Yes}, but with some caveats. Here’s a detailed assessment:

\textbf{Strengths}
\begin{itemize}
    \item \textbf{Versatile Removal:} Hits artifacts, enchantments, and tapped creatures — all common in Neon Dynasty, which has a lot of artifacts and enchantments.
    \item \textbf{Bonus Samurai Token:} If you control both an artifact and an enchantment, you get a 2/2 Vigilant Samurai, adding tempo and board presence.
    \item \textbf{Low Cost:} At just {W}{W}, it's efficient, though double-White can strain your mana if you're in multiple colors.
\end{itemize}

\textbf{Limitations}
\begin{itemize}
    \item \textbf{Sorcery Speed:} Can't be used as a combat trick or instant-speed interaction.
    \item \textbf{Conditional Token:} You need both an artifact and an enchantment in play to get the 2/2, which isn’t always guaranteed, especially early game.
    \item \textbf{Tapped Creature Clause:} You can only hit creatures that are tapped (likely after combat), so it doesn’t always work as proactive removal against untapped threats.
\end{itemize}

\textbf{When Should You Draft It?}\\
You want to draft \textbf{Banishing Slash} when:
\begin{enumerate}
    \item \textbf{You're in White (especially mono-White or heavy White decks).}
    \item \textbf{You're playing artifact and enchantment synergies,} which is very common in \textit{Neon Dynasty}:
    \begin{itemize}
        \item White often pairs with Blue or Green, both of which have plenty of enchantments and artifacts.
        \item Example archetypes:
        \begin{itemize}
            \item \textbf{White-Blue (Artifacts \& Vehicles)}
            \item \textbf{White-Green (Enchantments, ``Modified" synergy)}
        \end{itemize}
    \end{itemize}
    \item \textbf{Your deck lacks removal options,} and you need interaction for artifacts, enchantments, or evasive/tapped creatures.
    \item \textbf{You expect the bonus token is achievable,} improving the card’s value significantly.
\end{enumerate}

\textbf{Draft Pick Priority}
\begin{itemize}
    \item \textbf{High Priority} in:
    \begin{itemize}
        \item Heavy White decks
        \item White-based decks with strong artifact/enchantment support
    \end{itemize}
    \item \textbf{Medium Priority} if:
    \begin{itemize}
        \item You’re splashing White but struggling to meet the \textbf{{W}{W}} cost consistently
        \item Your deck doesn’t have much enchantment/artifact synergy
    \end{itemize}
    \item \textbf{Low Priority} if:
    \begin{itemize}
        \item You're not in White or your mana base makes \textbf{{W}{W}} awkward
    \end{itemize}
\end{itemize}

\textbf{Conclusion}\\
\textbf{Banishing Slash} is:
\begin{itemize}
    \item Solid removal
    \item Even better in synergistic decks
    \item More powerful in slower, grindier games where artifacts/enchantments are abundant
\end{itemize}

Would you like to run through some draft pick scenarios with this card? Or analyze another Neon Dynasty card?

\noindent\textbf{Llama-3-8B-Instruct:} I couldn't find any information on a card called ``Banishing Slash" from the Magic: The Gathering expansion ``Kamigawa: Neon Dynasty". It's possible that the card doesn't exist or is a non-foiling or promotional card.

However, I can suggest some general guidelines for evaluating a card's strength and drafting considerations:

\begin{enumerate}
    \item \textbf{Check the card's text:} Look at the card's effects, mana cost, and any relevant keywords (e.g., flying, trample, etc.).
    \item \textbf{Consider the card's rarity:} Rarer cards are often stronger and more desirable in a draft.
    \item \textbf{Assess the card's power level:} Evaluate the card's power level relative to other cards in the set. Is it a game-changer, or is it a solid but not spectacular card?
    \item \textbf{Think about the card's playstyle:} Does the card fit into a specific archetype or strategy? Is it a good fit for a particular color combination?
    \item \textbf{Consider the card's potential in different formats:} Will the card be strong in constructed formats, or is it more suited for casual or draft play?
\end{enumerate}

If you can provide more information about ``Banishing Slash" or the specific card you're interested in, I'd be happy to help you evaluate its strength and drafting considerations.

\bibliography{bib}

\end{document}